\documentclass[12pt]{l4dc2022}

\newcommand{\argmin}{\operatornamewithlimits{argmin}}

\usepackage{cleveref}

\title[ValueNetQP]{ValueNetQP: Learned one-step optimal control for legged locomotion}
\usepackage{times}
\usepackage{todonotes}

\author{%
 \Name{Julian Viereck} \Email{jviereck@nyu.edu}\\
 \AND
 \Name{Avadesh Meduri} \Email{am9789@nyu.edu}\\
 \AND
 \Name{Ludovic Righetti} \Email{ludovic.righetti@nyu.edu}\\
}

\begin{document}

\maketitle

\begin{abstract}%
Optimal control is a successful approach to generate motions for complex robots, in particular for legged locomotion. However, these techniques are often too slow to run in real time for model predictive control or one needs to drastically simplify the dynamics model. In this work, we present a method to learn to predict the gradient and hessian of the problem value function, enabling fast resolution of the predictive control problem with a one-step quadratic program. In addition, our method is able to satisfy constraints like friction cones and unilateral constraints, which are important for high dynamics locomotion tasks. We demonstrate the capability of our method in simulation and on a real quadruped robot performing trotting and bounding motions.
\end{abstract}


\section{Introduction}

Motion generation algorithms for legged robots can be broadly classified in two classes. On one side, there are optimal control methods~\cite{ponton2021efficient, shah2021rapid, carpentier2018multicontact, mastalli2020crocoddyl, winkler2018gait}. These methods are very general and versatile but are computationally demanding to run in realtime, for model-predictive control (MPC), on a robot. Oftentimes, optimal trajectories are computed offline first before being executed on the robot. On the other side are policy learning techniques like reinforcement learning~\cite{lee2020learning, peng2020learning, rudin2021learning, xie2020learning, tan2018sim}, which learn a control policy directly as a neural network. The policy is trained to optimize a long horizon reward. These methods are often sample inefficient, are prone to sim-to-real issues when transferring the simulated policy to the real robot and have a "hard-coded" policy as a neural network with limited constraint satisfaction guarantees.

To improve sample efficiency and solve times, model based optimal control algorithms have been used speed up policy learning. The Guided Policy Search algorithm~\cite{levine2013guided} uses an iterative linear quadratic regulator (iLQR) and optimizes trajectories together with a neural network policy till convergence. The work in~\cite{mordatch2014combining} also uses trajectory based optimization method and optimizes a neural network policy at the same time. However, these methods do not preserves all the information from the initial trajectory optimizer. In addition, there is no way to add constraints when running the policy.

Alternatively, learning a value function from iLQR has been explored. In~\cite{zhong2013value} the authors approximate the value function at the terminal step of a MPC problem using different function approximators. In~\cite{lowrey2018plan} the authors learn the value function and use MPC for control as well. The authors do not incorporate constraints when solving for the final control and also are limited by the speed of the MPC solver.

The idea to use an optimization algorithm with future reward prediction is part of the paper from ~\cite{kalashnikov2018qt}. There, the authors learn the Q function for a problem for visual manipulation. The optimal action to use on the robot is found by maximizing samples from the Q function given the current state. However, this method does not rely on a model of the dynamics.

Closer to our work, the authors in~\cite{parag2022value} propose a method to predict the value function gradient and hessian using Sobolev learning.
However, we were not able to get good learning results using Sobolev learning for our quadruped locomotion experiments. They also used a MPC approach with a longer horizon, which increases solve times but potentially improve robustness to the quality of learned value function. While they demonstrate promising results on simulated low dimensional problems, their method was not tested on systems with complexity comparable to a legged robot nor was it tested on real hardware.

In this paper, we propose a method to learn the gradient and Hessian of a value function originally computed by an optimal control solver (iLQR). We then formulate a simple quadratic program (QP) which efficiently uses these learned functions for one-step model predictive control at high control rates.
Our method enables to significantly decrease computation time while retaining the ability to generate the complex locomotion movements. By using the QP we are also able to enforce constraints on our solution like friction and unilateral contact force constraints, which is not possible with policy learning methods or with the iLQR algorithm. With this approach, we generate different dynamic locomotion gaits on a real quadruped robot such as trotting and bounding. We show that it is possible to robustly generate these dynamic motions, with a single step horizon, affording computation of optimal controls at 500 Hz.

Specifically, the paper contributions are as follows: 1) We propose a stable method for learning the value function gradient and Hessian information from iLQR, 2) we introduce our method for computing control commands from value function gradients and Hessians taking constraints into account, and 3) we demonstrate our method on two locomotion tasks on a real robot.
In the following sections we explain in detail our method. After this we describe our experimental setup and show results on the simulated and real robot. Finally, we conclude.

\section{Method}

\begin{figure}
  \centering
    \includegraphics[width=0.8\columnwidth]{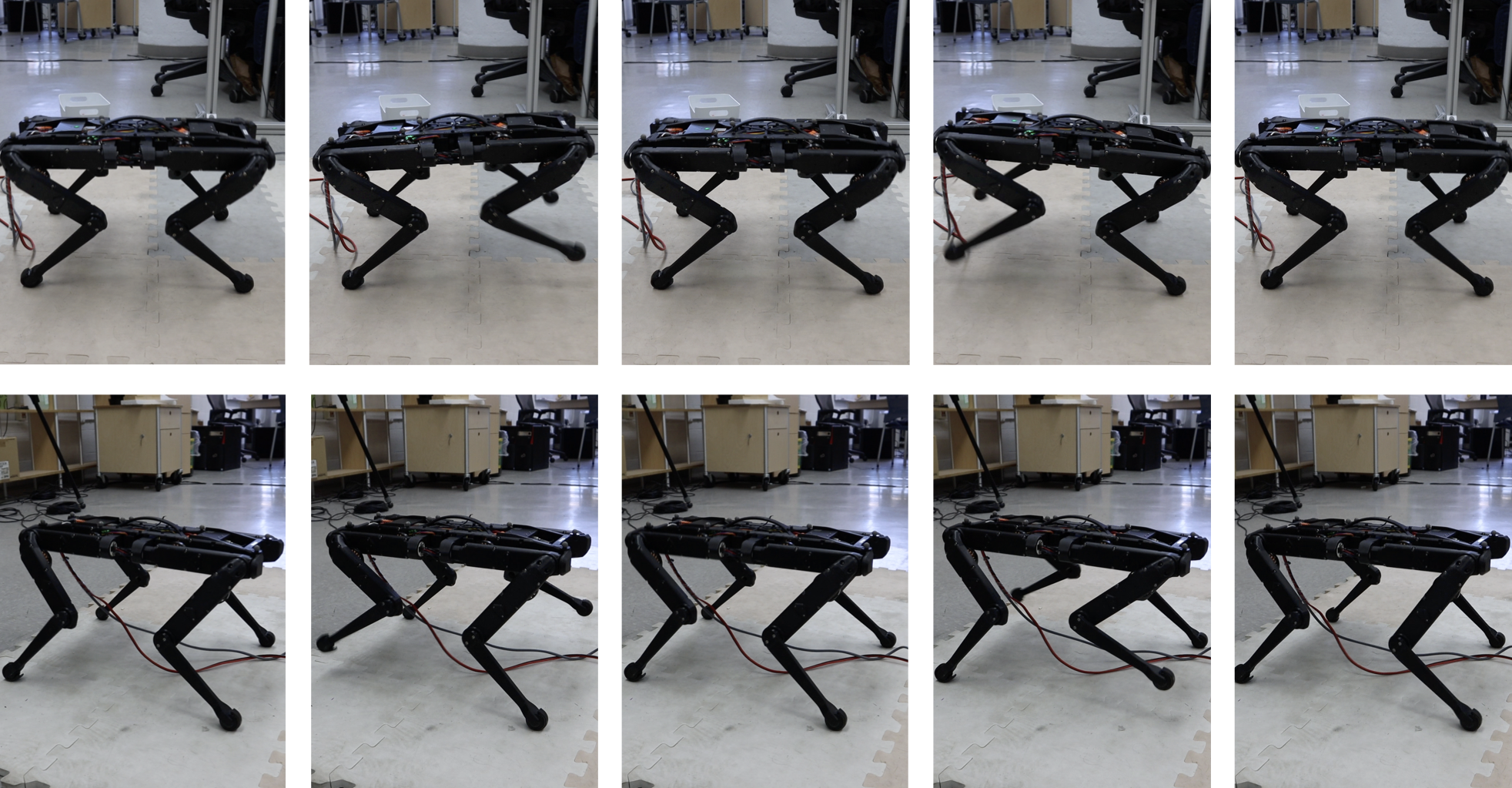}
  \caption{Time slices running the bounding (top) and trotting (bottom) motion on the real robot.}
  \label{fig:bound_trot}
\end{figure}

The method proposed in this paper is as follow. First, we optimize motions from different starting positions and with different properties (e.g. different desired velocity) using iLQR. Second, we use the optimized trajectories together with the associated value function expansion to learn a mapping between the current feature transformed state $\phi(x_t)$ to the value function gradient and hessian at the next time step. Finally, at run time, our optimal control problem is reduced to the resolution of a simple quadratic program (QP) which minimizes the learned value function for the state at the next time step while including important constraints such as friction pyramids and unilateral constraints.

The rest of this section presents the details of the approach. First, we re-state the iLQR recursions. Then we describe how control is optimized given a predicted value function using a QP. Finally, we explain how the value function gradients are learned using a neural network.

\subsection{Optimal control with iLQR}\label{sec:ilqr}
We aim to solve an optimal control problem of the form
\begin{align}
\label{eq:cost}
\min_{u_1, \dots, u_T}& \sum^{T}_{t = 1} l_t(x_t, u_t)\\
&\mathrm{s.t.} \ x_{t+1} = f_t(x_t, u_t) \nonumber
\end{align}
where $x_t$ and $u_t$ are the system state and control vectors, $l_t$ an instantaneous cost, and $f_t$ describes the system dynamics.
The iLQR algorithm computes a local optimal control law starting from a reference trajectory $\hat{x}$. The method takes a second order expansion of the loss function $l_t$ and a first order expansion of the dynamics $f_t$ around the reference trajectory. These approximations are used to compute backward Riccati equations. The algorithm works in two passes.

In the backwards pass, the algorithm computes the second order approximation of the value function $V$ and Q function along the reference trajectory and proposes a control update $\delta u$. The recursive update equations for the value function $V$ and $Q$ function are:

\begin{align}
    Q^x_t &= l^x_t + f^{x^T}_t V^x_{t+1}  \\
    Q^u_t &= l^u_t + f^{u^T}_t V^x_{t+1}  \\
    Q^{xx}_t &= l^{xx}_t + f^{x^T}_t V^{xx}_{t+1} f^{x}_t + V^{x}_{t+1} f^{xx}  \\
    Q^{ux}_t &= l^{ux}_t + f^{u^T}_t V^{xx}_{t+1} f^{x}_t + V^{x}_{t+1} f^{ux}  \\
    Q^{uu}_t &= l^{uu}_t + f^{u^T}_t V^{xx}_{t+1} f^{u}_t + V^{x}_{t+1} f^{uu}  \\
    V^x_t &= Q^{x}_t - Q^{xu}_t \left(Q_t^{uu}\right)^{-1} Q^u_t \\
    V^{xx}_t &= Q^{xx}_t - Q^{xu}_t \left(Q_t^{uu}\right)^{-1} Q_t^{ux},
\end{align}
where $\cdot_t^x, \cdot_t^u$ denotes the gradient at time $t$ with respect to $x$ and $u$ and $\cdot^{xx}_t, \cdot^{ux}_t, \cdot^{uu}_t$ denotes the Hessian terms with respect to $xx, ux$ and $uu$ at time $t$.

In the forward pass, a line search is used to determine the best update $\alpha \delta u$ that would reduce the trajectory cost when integrating the dynamics forward.
To compute the control update $\delta u$, we use the BoxDDP algorithm (\cite{tassa2014control}). This allows us to put bounds on the control and e.g. avoid negative forces in the vertical direction.

\subsection{Solving for control using a QP}
From Bellman's principle of optimally, it follows that
to compute the optimal control $u_t$ and value function at time $t$, it is sufficient to know the current cost and value function at time $t+1$:
\begin{align}
    V(x_t) &= \min_{u_t}[l_t(x_t, u_t) + V(f_t(x_t, u_t))].
\end{align}
Our goal is then to reduce the original optimal control problem to a one-step optimization problem (as a QP) that will leverage a learned value function gradient and Hessian.
We write this equation as a constrained optimization problem:
\begin{align}
    V(x_t) = & \min_{u_t, x_{t+1}}[l_t(x_t, u_t) + V(x_{t+1})].\\
         &\text{subject to } x_{t+1} = f_t(x_t, u_t) \nonumber
\end{align}
Recall that in iLQR the value function is expressed along a nominal trajectory $\hat{x}_t$. We denote the second order approximation for the value function as $V(x_{t+1}) \approx V^0_{t+1} + V^x_{t+1} (x_{t+1} - \hat{x}_{t+1}) + \frac{1}{2} (x_{t+1} - \hat{x}_{t+1})^T V^{xx}_{t+1} (x_{t+1} - \hat{x}_{t+1})$. Assuming a cost term of the form $l_t(x_t, u_t) = l_t(x_t) + \frac{1}{2} u_t^T R^{uu}_{t} u_t$ we can solve for the arguments in the last minimization problem as follows:
\begin{align}
    \argmin_{u_t, x_{t+1}}& \left[
        l_t(x_t) + \frac{1}{2} u_t^T R^{uu}_{t} u_t +
        V^0_{t+1} + V^x_{t+1} (x_{t+1} - \hat{x}_{t + 1}) + \frac{1}{2} (x_{t + 1} - \hat{x}_{t+1})^T V^{xx}_{t+1} (x_{t + 1} - \hat{x}_{t+1}) \right]\nonumber \\
    & \quad \text{subject to}\quad x_{t+1} = f(x_t, u_t) \nonumber
\end{align}
which is equivalent to
\begin{align}
    \argmin_{u_t, x_{t+1}}& \left[
        (V^x_{t+1} - \hat{x}_{t+1}^T V^{xx}_{t+1})x_{t+1} + \frac{1}{2} (x_{t + 1}^T V^{xx}_{t+1} x_{t + 1} + u_t^T R^{uu}_{t} u_t)
    \right], \label{eq:min_qp_full}\\
    &\text{subject to}\quad x_{t+1} = f(x_t, u_t) \nonumber
\end{align}
where we removed constant terms and used the fact that $V^{xx}_{t+1}$ is symmetric. Interestingly, our problem only involves the gradient and Hessian of the value function and not its actual value. Also, since our formulation contains a one step look up into the horizon it makes the nonlinear dynamics $f(x,u)$ linear. Subsequently, we can efficiently solve the above optimization problem with a QP to obtain the optimal control $u_{t}$.

\subsection{Using the dynamics structure to reduce the QP}
Since we work with system with a second order dynamics, we can separate the state $x_t$ into the position $s_t$ and velocity $v_t$ part $x_t = \begin{bmatrix}s_t, v_t\end{bmatrix}$. We then explicitly include the numerical integration scheme to simplify the previous QP. This will be important to facilitate learning.
Starting from~\cref{eq:min_qp_full}, writing the block components of $V^{xx}_{t+1}$ corresponding to $s_t$ and $v_t$ as $V^{xx}_{t+1} = \begin{bmatrix}V^{ss}_{t+1} & V^{sv}_{t+1} \\ V^{vs}_{t+1} & V^{vv}_{t+1}\end{bmatrix}$ and $(V^x_{t+1} - \hat{x}_{t+1}^T V^{xx}_{t+1}) = \begin{bmatrix} V^s_{t+1} \\ V^v_{t+1} \end{bmatrix}$ we get
\begin{align}
    \argmin_{u_t, v_{t+1}}& \left[
        \begin{bmatrix} V^s_{t+1} \\ V^v_{t+1} \end{bmatrix}^T\begin{bmatrix}s_{t+1}\\v_{t+1}\end{bmatrix} + \frac{1}{2} \left(\begin{bmatrix}s_{t+1}\\v_{t+1}\end{bmatrix}^T \begin{bmatrix}V^{ss}_{t+1} & V^{sv}_{t+1} \\ V^{vs}_{t+1} & V^{vv}_{t+1}\end{bmatrix} \begin{bmatrix}s_{t+1}\\v_{t+1}\end{bmatrix} + u_t^T R^{uu}_{t} u_t \right)
    \right], \\
    &\text{subject to } \quad \begin{bmatrix}s_{t+1}\\v_{t+1}\end{bmatrix} = \begin{bmatrix}s_{t} + \Delta t~v_t \\ v_t + B^u u_t + B^0\end{bmatrix} \nonumber
\end{align}
where we rewrote the constraints on the dynamics using an Euler integration scheme $s_{t+1} = s_{t} + \Delta t~v_t$ and linearization of the velocity dynamics $B^u$ and offset $B^0$. Because $s_{t+1}$ is constrained by $s_t$ and $v_t$, which are not part of the optimization variables, $s_{t+1}$ can be replaced in the last optimization with $s_{t} + \Delta t~v_t$. Applying this substitution and simplifying terms yields:

\begin{align}
    u_t^*, v_{t+1}^* = &\argmin_{u_t, v_{t+1}} \left[
        (V^v_{t+1} + s_{t+1} V^{sv}_{t+1}) v_{t+1} + \frac{1}{2} \left( v_{t+1} V^{vv}_{t+1} v_{t+1} + u_t^T R^{uu}_{t} u_t\right)
    \right]. \label{eq:min_qp_min} \\
    &\text{subject to } v_{t+1} = v_t + B^u u_t + B^0 \nonumber
\end{align}
This formulation of the problem has multiple important benefits. First, the dimension of the optimization problem is smaller. In addition, the vector $(V^v_{t+1} + x_{t+1} V^{sv}_{t+1})$ and the matrix $V^{vv}_{t+1}$ are smaller compared to equivalent entities written in the original problem. This size reduction will be very important to yield good results once we learn these quantities using a neural network. Finally, it will be easy to add state and control
constraints to this QP, as we will show in the subsequent sections.

\subsection{Learning the value function gradient and Hessian}
Given optimized iLQR trajectories, we can also recover the value function gradient and Hessian at each time steps (cf. recursions of Sec. \ref{sec:ilqr}). We aim to predict the value function gradient $g_{t+1} = (V^v_{t+1} + s_{t+1} V^{sv}_{t+1})$ and Hessian $V^{vv}_{t+1}$. We do this by regressing the gradient and Hessian directly to a feature transformed input state $\phi(x_t)$ (we will give an example of a feature transformation of the input state in the experimental section below). The regression is solved under a L1 loss using gradient descent and a neural network. We call the neural network the value function network (ValueNet). As for the network output, we flatten the Hessian matrix $V^{vv}_{t+1}$ into a vector and concatenate it with the gradient $g$ into a single target vector for the regression task.
To use the Hessian $V^{vv}_{t+1}$ in the QP we must ensure that the matrix is positive definite. We do this as follows: First, we make sure the matrix is symmetric by instead computing $1/2 \left(V^{vv}_{t+1} + V^{{vv}^T}_{t+1} \right)$. Then, we compute the eigenvalue and eigenvector of the matrix and set negative eigenvalues to small positive eigenvalues. We found in our experimental results that this regularization worked well.

In all our experiments, we use the same neural network architecture. We use a 3 layer fully connected feedforward neural network with 256 neurons per layer. As activation function we use tanh. The network is trained for 256 epochs with a batch size of 128 using Adam and a learning rate of 3e-4.
During our experiments we noticed that small changes in the predicted gradient and Hessian lead to drastic different results when solving the QP. To mediate this problem, we found it useful to normalize the gradient and hessian prediction to zero mean and unit variance.

\section{Experimental setup}
In the following, we describe our experimental setup and experiments. We demonstrate our method on two kinds of locomotion tasks: bounding and trotting on a simulated and real quadruped robot, where the desired velocity can be controlled and the robot can handle push perturbations. In our experiments we use the Solo12 quadruped from the Open Dynamic Robot Initiative (\cite{grimminger2020open}). The 12 joints are torque controlled, making it an ideal platform to test model predictive control methods. In all experiments, we use a Vicon motion capture system and IMU to get an estimate of the robot base position and velocity.

\subsection{Dynamics model}
To illustrate the capabilities of our approach, we use
a standard, simplified centroidal dynamics model of a quadruped. We model the state $x_t$ as $x_t = [c_t, \alpha_t, \dot{c}_t, \omega]$ where $c_t$ is the position of the center of mass (CoM), $\alpha_t$ is the orientation of the base and $\dot{c}_t, \omega$ are the respective velocities. All quantities are expressed in an inertial (world) frame. The control inputs are the forces applied by the feet of the quadruped $F_i$,  where $F_i$ denotes the contact force at the $i$-th leg and $r_i$ the position of the $i$-th leg contact location. The discretized dynamics equations are then
\begin{align}
    \label{eq:dyn}
    \begin{bmatrix}
        c_{t + 1} \\
        \alpha_{t + 1} \\
        \dot{c}_{t + 1} \\
        \omega_{t + 1}
    \end{bmatrix} =
    \begin{bmatrix}
        c_{t} \\
        \alpha_{t} \\
        \dot{c}_{t} \\
        \omega_{t}
    \end{bmatrix} + \Delta t  \begin{bmatrix}
        \dot{c}_t \\
        \omega_t \\
        \sum_i \frac{F_i}{m} - G\\
        \sum_i \frac{F_i \times (r_i - c_t)}{I}
    \end{bmatrix},
\end{align}
where $G = [0, 0, 9.81]^T~m/s^2$ is the gravity vector and $m= 2.5~kg$ is the robot mass. $I$ denotes the rotational inertia mass matrix of the base joint. Since the inertia mass matrix does not change significantly for trotting and bounding, we keep it constant throughout all of our experiments. Note however that the associated optimal control problem is not convex due to the cross product.

\subsection{Pattern generator}
To generate example bounding and trotting motions, we need to know the desired foot locations $r_i$. For this, we utilize a pattern generator. Given an initial CoM position and velocity, desired CoM velocity and information about the gait (i.e. the sequence of endeffector contacts with the ground) , the pattern generator generates the foot locations $r_i$.

When the foot $i$ goes into contact, the foot location is computed using the following Raibert-inspired heuristic~\cite{raibert1984experiments}
\begin{align}
    \label{eq:foot}
    r_i = c_t + \text{shoulder}_i + \frac{t_\text{stance}}{2}  \dot{c} + k_\text{raibert} (v_\text{cmd} - \dot{c}),
\end{align}
where $\text{shoulder}_i$ is the offset of the endeffector in the neutral position from the CoM, $t_\text{stance}$ is stance  phase duration, $k_\text{raibert}$ is a constant (we choose 0.03 in all our experiments) and $v_\text{cmd}$ is the commanded / desired velocity of the CoM.
To evaluate this equation, we
interpolate between the current CoM velocity $\dot{c}$ and the desired velocity. In particular, at each timestep we update the planned CoM velocity as
\begin{align}
    \dot{c} = (1 - v_\alpha) ~ \dot{c} + v_\alpha ~ v_\text{cmd},
\end{align}
where we choose $v_\alpha$ as 0.02 in all our experiments.
Using this control, the robot will bring the base velocity slowly towards the desired velocity.

\subsection{iLQR data generation and network training}

To train our ValueNet we use samples from iLQR optimized trajectories. We generate many iLQR trajectories by putting the robot into a random initial state (random initial position and velocity), sample a desired velocity command in a random direction (up to 0.6 m/s) and generate the feet locations $r_i$ using the pattern generator. As iLQR cost $l(x_t, u_t)$ in~\cref{eq:cost} we use a quadratic cost between the current state $x_t$ and desired state $\hat{x}_t$ where there is no weight for the horizontal position. The desired state is $\hat{x} = [0, 0, 0.21, v^x_{cmd}, v^y_{cmd}, 0, \dots, 0]$. In addition, we use a quadratic cost to regularize the control.

Given that we optimize a finite horizon problem, we know that the value function gradient and Hessian will be very different from their stationary values for the infinite horizon problem towards the end of the horizon.  This change in magnitude makes learning to predict the value function gradient and Hessian difficult if value function information at the end of the optimized trajectory was used. To overcome this problem, we optimize many small iLQR trajectories and use only the information from the first timestep. We do this as follows: We first optimize a longer (four gait cycles long) horizon trajectory from a random initial configuration using iLQR. Then, for each of the timesteps belonging to the first 1.5 cycles along this trajectory, we initialize a shorter iLQR trajectory (2.5 gait cycles long). These shorter iLQR trajectories are warm started using the longer iLQR trajectories and optimized. We then store the information from the first time step in form of $\phi(x_0)$ as well as value function gradient and Hessian. For the trotting and bounding motion we optimize 2048 long iLQR trajectories leading to around 313000 training samples. As time discretization we use $\Delta t = 0.004~s$.


As discussed before, we perform a feature transformation $\phi(x_t)$ on our states before using them as input to the neural network. It is important that the network does not overfit to the current absolute position of the robot in the horizontal plane. Therefore, $\phi(x_t)$ is using the full state $x_t$ besides the CoM position in horizontal plane. Besides this reduced state, we found it beneficial to include also the relative positions of the endeffectors with respect to the CoM in $\phi(x_t)$. We also incorporate information whether an endeffector is in contact with the ground at a given state. We do this by encoding the active/inactive contact state as \{0, 1\} respectively and pass four numbers (one for each leg) as input. Lastly, we also pass the desired velocity $v_\text{cmd}$ and the contact time of each leg as input. The contact time is the time until the foot either is about to leave contact or how long it is still in contact (depending on the current contact configuration).

\subsection{Integration of QP in Whole body control}
At each control cycle, forces $F_i$ are computed by specializing the QP in~\cref{eq:min_qp_min} with the dynamic models from~\cref{eq:dyn}, adding important contact constraints and the learned ValueNet:
\begin{align}
    F_t^*, v_{t+1}^* = &\argmin_{F_t, v_{t+1}} \left[
        g_{t+1} v_{t+1} + \frac{1}{2} v_{t+1} V^{vv}_{t+1} v_{t+1} + \frac{1}{2} F_t^T R^{FF}_{t} F_t
    \right]. \label{eq:min_qp_min_final} \\
    &\text{subject to }
        \begin{matrix}
            v_{t+1} = v_t + B^F F_t + B^0 \\
            |F_x| \leq \mu F_z \\
            |F_y| \leq \mu F_z \\
            0 \leq F_z \leq 30,
        \end{matrix} \nonumber
\end{align}
where we added friction pyramid constraints. We use a friction coefficient of $\mu = 0.6$ in our experiments. We also impose unilateral contact forces and limit the z-force to $30~Nm$ for safety reason (roughly double of what is usually applied). If not stated otherwise, the final controller runs at 500 Hz.

We use a whole-body controller to map these commands to actuation torques. First, we must generate motions for the endeffector to reach the desired location $r_i$. Using~\cref{eq:foot} we compute the desired location of the swing-feet and plan a fifth order polynominal trajectory for the x/y direction and 3rd order polynominal for the z direction in Cartesian space for the endeffector motions. We track the desired endeffector position $\hat{e}_i$ and velocity $\dot{\hat{e}}_i$ using the following impedance controller
\begin{align}
    \tau = \sum_i J^T_{c,i} \left\{P (\hat{e}_i - e_i) + D (\dot{\hat{e}}_i - \dot{e}_i) + F_i\right\}, \label{eq:torque}
\end{align}
where $\tau$ is vector of actuation torques, $J_{c,i}$ is the contact Jacobian for leg $i$, $P$ and $D$ are gains and $e_i$ is the position of the endeffector. It is important, however, to keep in mind that the most important part of the locomotion behavior, including proper balancing and CoM motion, is generated by the one-step QP solver.

\begin{figure}
  \centering
    \includegraphics[width=1.0\columnwidth]{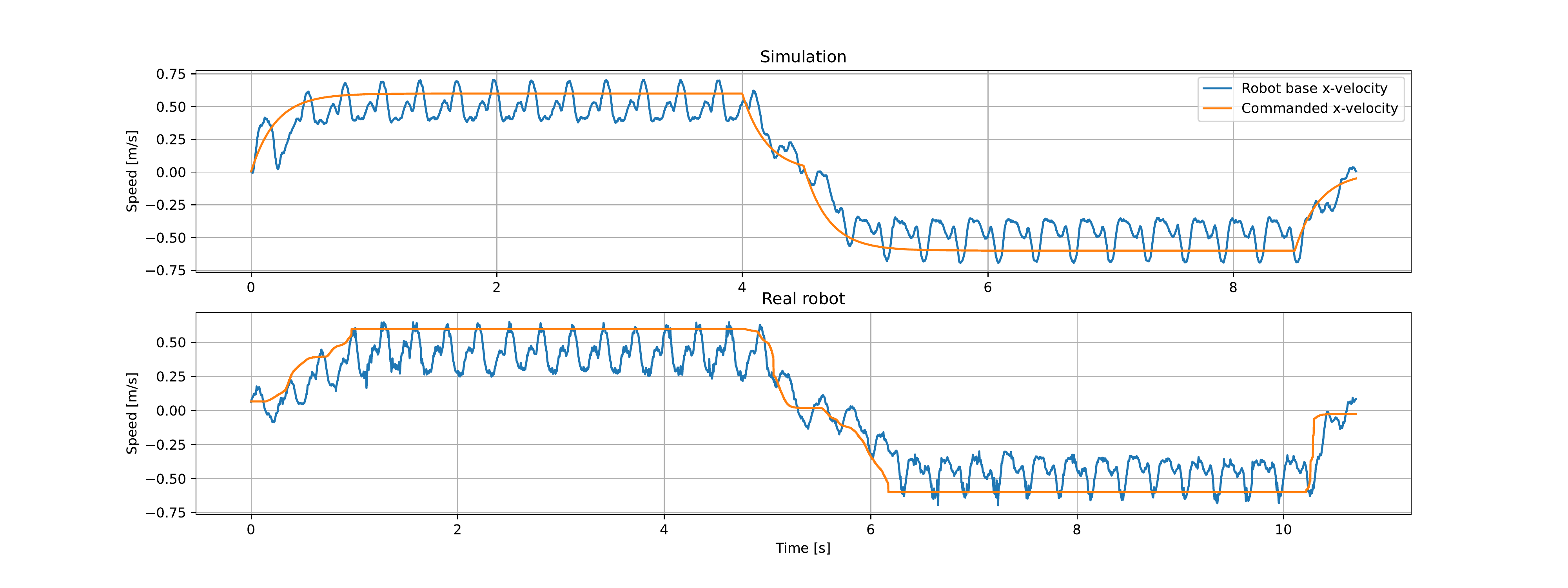}
  \caption{Velocity tracking on the simulated (top) and real robot (bottom) for a bounding motion.}
  \label{fig:plot_vel_bound_tracking}
\end{figure}

\begin{figure}
  \centering
    \includegraphics[width=1.0\columnwidth]{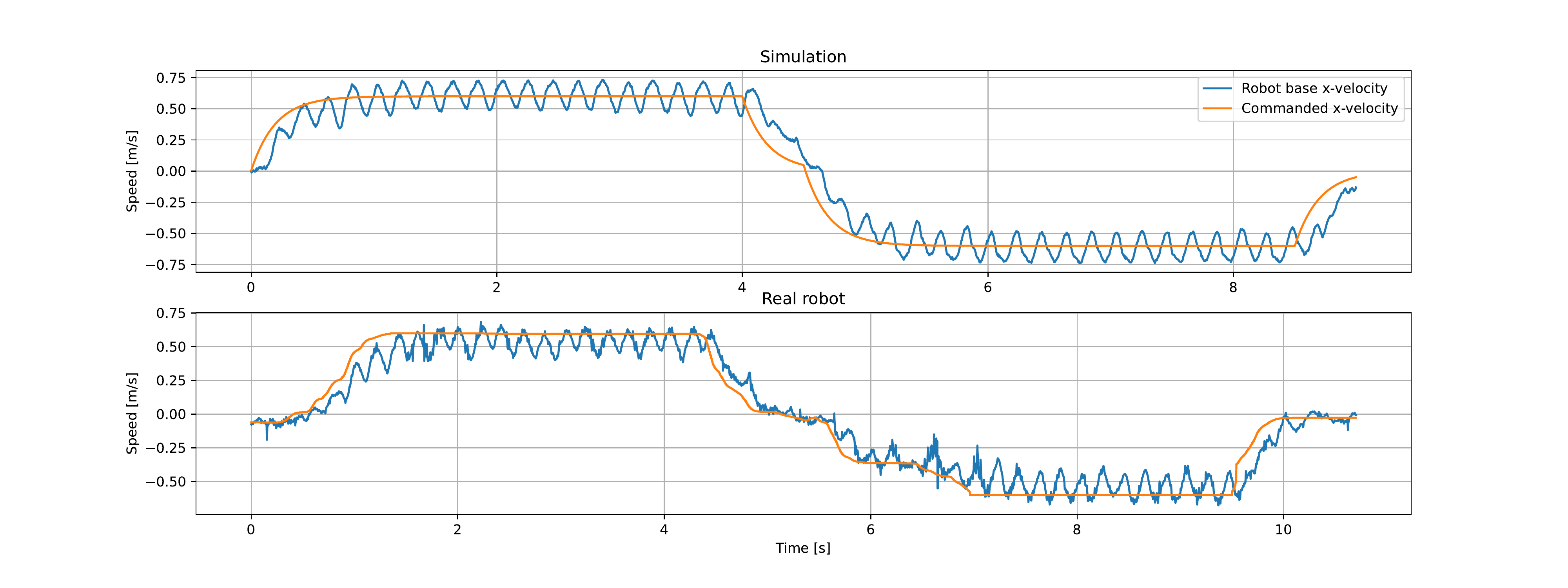}
  \caption{Velocity tracking on the simulated (top) and real robot (bottom) for a trotting motion.}
  \label{fig:plot_vel_trot_tracking}
\end{figure}
\section{Experiments and results}

Using the method and experimental setup described above, we perform a set of experiments in simulation and on the real hardware. The experiments are designed to demonstrate that our proposed method works and can be robustly applied on a real robot. Examples of the motions can be found in~\cref{fig:bound_trot} and in the accompanying video \footnote{Video of the experimental results \url{https://youtu.be/qcBwlyZjnRA}}.
\subsection{Velocity tracking}
\begin{figure}
  \centering
    \includegraphics[width=1.0\columnwidth]{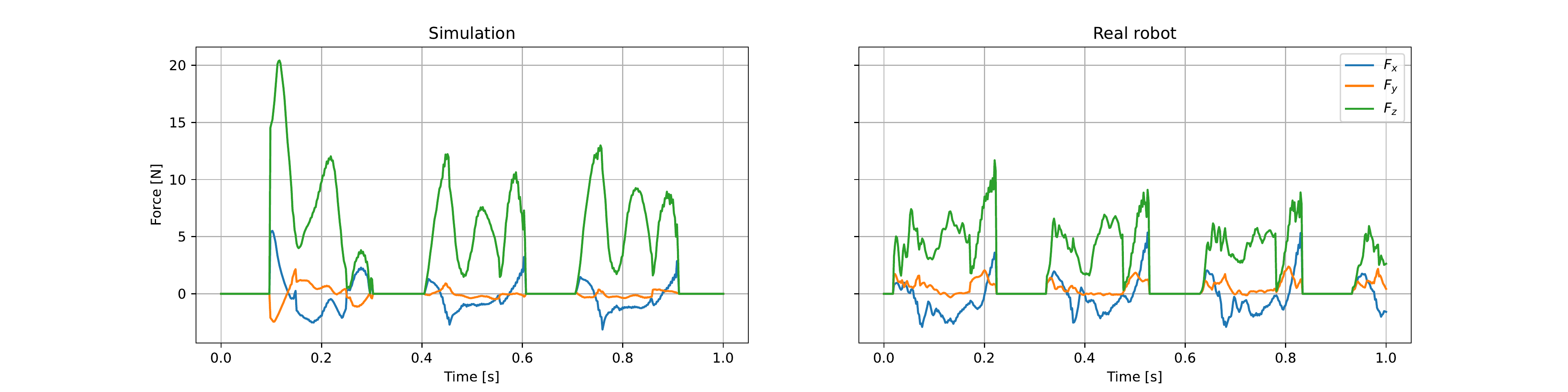}
  \caption{Force profile of the first second from tracking the bounding motion in~\cref{fig:plot_vel_bound_tracking}. The left plot shows the results from simulation and the right plot shows results from the real robot.}
  \label{fig:plot_vel_bound_forces}
\end{figure}

We evaluate the ability of our method to track a desired velocity, using the bounding and trotting motion on the simulated and real robot. We let the robot bound/trot forward and backwards for a few seconds at different commanded velocities. The results are shown in~\cref{fig:plot_vel_bound_tracking} and~\cref{fig:plot_vel_trot_tracking}.
We observe that the velocity is tracked very well both in simulation and on the real robot, while the exact same controller and learned value function are used.
It is interesting to see that the one-step QP is able to generate a diverse set of stable motions thanks to the learned value function.


\subsection{Robustness to external disturbances}

To test robustness of our proposed method, we perform two sets of experiments on the real robot: First, we manually push the robot while keeping it in place. In the second experiment, we place obstacles with 2.5~cm height in the path of travel and let the robot run across these obstacles. We repeat both experiments for the trotting and bounding motion. The results are shown in the accompanying video.
As one can see in the video, our method is able to keep the robot stable when moving over the obstacles and is robust to random pushes.

\subsection{Smoothness of control profile}

Recall that our method is computing desired forces at the endeffectors as control output. Especially when working on the real hardware it is important that the computed forces (and thereby the computed torques using~\cref{eq:torque}) are smooth enough to prevent the robot from shaking and destabilizing. In~\cref{fig:plot_vel_bound_forces}, we show the force profile generated by our method on the bounding task shown in~\cref{fig:plot_vel_bound_tracking} for the first second.

As one can see from the plot, the force profile is quite smooth in simulation. On the real robot, the forces are a bit more noisy but this is expected given noisy measurements from the real robot that are fedback in the QP (our simulation is noise free). Still, the forces on the real system are smooth enough and not destabilize the bounding or trotting motion.

\subsection{Necessity for QP constraints}

One of the benefits of our method is that we are able to add constraints when solving for the control in~\cref{eq:min_qp_min}. We verify the necessity for the bounds by disabling them for the trotting and bounding motion.
When we disable the QP bounds for the trotting motion, we see negative $F_z$ forces being applied. Still, the motion of the robot stays robust. In contrast, when disabling the QP bounds on the bounding motion, the legs start slipping and the robot motion becomes unstable. This demonstrates the necessity of the QP bounds and the benefit of having these constraints.

\subsection{Reducing value function gradient and hessian prediction frequency}

By default we are evaluating the neural network to get a new value function gradient and Hessian at each control step at 500 Hz. In this experiment we study the prediction frequency has on the stability of the motion on the robot. To do this,  we reduce the update frequency of the predicted gradient and Hessian while running the rest of the controller (evaluating the QP and impedance controller) at 500~Hz. When running the bounding motion, we are able to reduce the value function prediction from 500~Hz to 62.5~Hz before the robot becomes unstable (shown in the accompanying video). This is an interesting observation, as it shows that it is not necessary to evaluate the value function at the same rate as the controller, therefore reducing computation.

\section{Conclusion and future work}

In this paper, we presented a method to reduce a (non-convex) optimal control problem into a one-step QP by learning the value function gradients and Hessians computed on the original problem. We demonstrated the approach on
 trotting and bounding motions with velocity control and showed that the method could be directly used on a real quadruped robot. This approach enables to significantly reduce computational complexity, enabling model predictive control to generate non-trivial locomotion behaviors.
 We demonstrated and analyzed the robustness and capabilities of this method in simulation and on a real quadruped robot.
As future work we are planning to study how the learned value function gradient and Hessian information can be used as terminal costs for longer horizon model predictive control. In addition, we intend to apply the method on a biped robot, which is more unstable and a humanoid robot with more degrees of freedom.

\acks{This work was supported by New York University, the European Union's Horizon 2020 research and innovation program (grant agreement 780684) and the National Science Foundation (grants 1825993, 1932187, 1925079 and 2026479).
}

\typeout{}
\bibliography{references}

\end{document}